\pdfoutput=1

\documentclass[11pt]{article}

\usepackage{ACL2023}

\usepackage{times}
\usepackage{latexsym}

\usepackage[T1]{fontenc}

\usepackage[utf8]{inputenc}

\usepackage{microtype}

\usepackage{inconsolata}

\usepackage{pgfplots}
\usepackage{pgfplotstable}
\usepgfplotslibrary{colorbrewer}
\usetikzlibrary{pgfplots.statistics, pgfplots.colorbrewer}
\pgfplotsset{compat = 1.15, 
             cycle list/Set1,
        }

\usepackage{caption}
\usepackage{subcaption}

\title{Efficient Black-Box Adversarial Attacks on Neural Text Detectors}

\author{Vitalii Fishchuk\\
University of Twente\\
  Faculty of Electrical Engineering,\\
  Mathematics and Computer Science\\
  \texttt{v.fishchuk@student.utwente.nl} \\\And
  Daniel Braun \\
  University of Twente \\
  Department of High-tech Business\\
  and Entrepreneurship\\
  \texttt{d.braun@utwente.nl}
  \\}

\begin{document}
\maketitle

\begin{abstract}
Neural text detectors are models trained to detect whether a given text was generated by a language model or written by a human. In this paper, we investigate three simple and resource-efficient strategies (parameter tweaking, prompt engineering, and character-level mutations) to alter texts generated by GPT-3.5 that are unsuspicious or unnoticeable for humans but cause misclassification by neural text detectors. The results show that especially parameter tweaking and character-level mutations are effective strategies.
\end{abstract}

\section{Introduction}
\label{sec:intro}
The widespread availability of neural text generation models, like ChatGPT, has caused an increased desire for neural text detectors, i.e. models that can detect whether a given text was AI-generated. The reliance of, e.g., educational institutions on such detectors has raised questions about their robustness in general and in specific with regard to adversarial attacks \citep{jawahar-etal-2020-automatic,wolff2022attacking,liang2023mutationbased,electronics12081948}. Such attacks exploit the fact that machine learning models by identifying patterns in the data rather than by understanding actual underlying concepts. Consequently, introducing small, human-unnoticeable perturbations can result in misclassification. \citep{Goodfellow2014, Szegedy2013}

Adversarial attacks can be categorised into black-box and white-box attacks \cite{Peng2023}. In white-box attacks, the attacker has full access to the target model, including its parameters, architecture, and loss function \cite{ebrahimi-etal-2018-hotflip, Gao2018}. During black-box attacks, the adversary can only input queries and observe the outputs without any insights into internal processing \cite{Gao2018}. Furthermore, it can be distinguished between targeted and untargeted attacks, where targeted attacks aim at triggering misclassification towards a specific label, while untargeted aim to cause any misclassification \cite{Rathore2021}.

This paper investigates effective and resource-efficient universal attack strategies in a black-box scenario with minimal resources, based on text generated with GPT 3.5 and three neural text detectors: the widely used open source GPT-2 Output detector model\footnote{\url{https://github.com/openai/gpt-2-output-dataset/tree/master/detector}}, the OpenAI text classifier\footnote{\url{https://platform.openai.com/ai-text-classifier}}, and the commercial Turnitin AI detector\footnote{\url{https://www.turnitin.com}}, which is used by many educational institutions. The results show that character-level mutations, tweaking the parameters of the generative model, as well as prompt engineering, are efficient and effective strategies, showing that currently available neural text detectors can not reliably detect texts generated by state-of-the-art large language models (LLMs).

\section{Related work}
Most of the existing literature about adversarial attacks focuses on image detection. Textual input is less used due to its discrete nature and the difficulty in introducing human-imperceptible perturbations, contrary to the image data, where a change in a few hundred pixels can go unnoticed  \cite{Jin2019, Peng2023}. Examples of adversarial attacks on general text classification models include the work by \citet{ebrahimi-etal-2018-hotflip} and \citet{Gao2018}. More recent work has started to specifically look into adversarial attacks on neural text detectors: \citet{wolff2022attacking} showed that introducing spelling mistakes and replacing characters with homoglyphs can significantly reduce the detection rate for GPT-2 texts. \citet{liang2023mutationbased} showed that similar character-level mutation-based attacks are also successful for RoBERTa-based detection models. \citet{Liang2023} not only showed that existing detectors are vulnerable to simple rephrasing, but they also showed that they are biased towards flagging texts that have been (manually) written by non-native speakers as AI-generated.

Because currently available methods are vulnerable to adversarial attacks, multiple suggestions have been made to improve their robustness, e.g. by \citet{electronics12081948}, \citet{Shen2023}, \citet{Crothers2022}, and \citet{Yoo2022}. While watermarking techniques to identify AI-generated texts are also investigated, they are generally seen as vulnerable to adversarial attacks, especially to mutation and paraphrasing-based approaches \citep{Jin2019,Kirchenbauer2023,Sadasivan2023}.

\section{Approaches}
\label{sec:approaches}
Based on the existing literature, we identified three promising and efficient approaches for adversarial attacks: parameter tweaking, prompt engineering and character-level mutations. All approaches were tested with the three neural text detectors mentioned in Section \ref{sec:intro}: GPT-2 output detector, OpenAI classifier, and Turnitin AI writing detector. The basis for all attacks were texts generated by the GPT-3.5-turbo model via the OpenAI API. The text samples were produced as 500-word essays, with topics taken from a list of 200 essay topics \cite{essayTopics} and the prompt \textit{``Write a five-hundred-word argumentative essay on the topic ‘topic’.''}. It was then evaluated how the detection rate changed between the original texts and their altered version. To ensure comparability of the results between different detectors, all scores were projected onto a scale from 0.0 (very likely not AI-generated) to 1.0 (very likely AI-generated). The GPT-2 Output detector returns a score between 0.0 and 1.0, that can be used directly. Turnitin returns a percentage between 0 and 100 indicating how much of the text was generated by AI. We divide the score by 100. The OpenAI classifier returns one of five labels (``very unlikely'', ``unlikely'', ``unclear'', ``possibly'', ``likely''). For each of the labels, \citet{openaiClassifier} provides a corresponding range of numerical thresholds, of which we take the mean score (0.05, 0.275, 0.675, 0.94, 0.99). The code for the evaluation was written with the assistance of GPT-4, followed by extensive testing of the code, as well as additional, manually implemented, features. The code and the data are available on GitHub\footnote{\url{https://github.com/Lolya-cloud/adversarial-attacks-on-neural-text-detectors}}.

\begin{table}[t]
    \centering
    \begin{tabular}{l r r r}
    \hline
        Parameter & Min & Max & Default \\\hline        
        Temperature & 0.0 & 2.0 & 1.0\\
        Top p & 0.0 & 1.0 & 1.0\\
        Frequency penalty & -2.0 & 2.0 & 0.0\\
        Presence penalty& -2.0 & 2.0 & 0.0\\\hline
    \end{tabular}
    \caption{Investigated parameters\vspace{-0.4cm}}
    \label{tab:param}
\end{table}

\begin{figure*}[t!]
    \centering
\begin{subfigure}[b]{0.45\textwidth}
    \centering
    \begin{tikzpicture}
        \begin{axis}[
            height=5cm,
            width=\linewidth,
            ylabel = {\small detection score},
            legend style={nodes={scale=0.8, transform shape}},
            legend pos=north east]
        \addplot +[mark=square] table [x=Frequency penalty, y=Turnitin, col sep=comma] {data/frequency_penalty_combined_scores.csv};
        \addplot +[mark=o] table [x=Frequency penalty, y=OpenAI classifier, col sep=comma] {data/frequency_penalty_combined_scores.csv};
        \addplot +[mark=triangle] table [x=Frequency penalty, y=GPT2 detector, col sep=comma] {data/frequency_penalty_combined_scores.csv};
        \addlegendentry{Turnitin}
        \addlegendentry{OpenAI}
        \addlegendentry{GPT2 detector}
        \end{axis}
    \end{tikzpicture}
    \caption{frequency penalty}
    \label{fig:frequency}
\end{subfigure}
\begin{subfigure}[b]{0.45\textwidth}
    \centering
    \begin{tikzpicture}
        \begin{axis}[
            height=5cm,
            width=\linewidth,
            legend pos=north east]
        \addplot +[mark=square] table [x=Presence penalty, y=Turnitin, col sep=comma] {data/presence_penalty_combined_scores.csv};
        \addplot +[mark=o] table [x=Presence penalty, y=OpenAI classifier, col sep=comma] {data/presence_penalty_combined_scores.csv};
        \addplot  +[mark=triangle] table [x=Presence penalty, y=GPT2 detector, col sep=comma] {data/presence_penalty_combined_scores.csv};
        \end{axis}
    \end{tikzpicture}
    \caption{presence penalty}
    \label{fig:presence}
\end{subfigure}
    \caption{Influence of the frequency and presence penalty on the detection score\label{fig:influence}}    
\end{figure*}
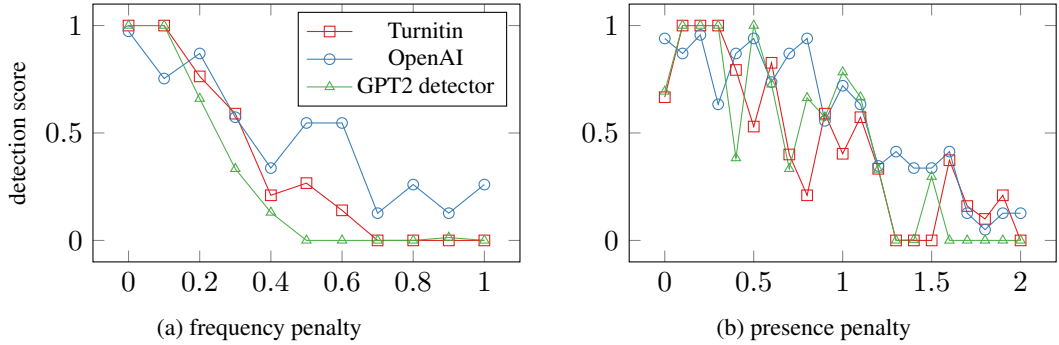
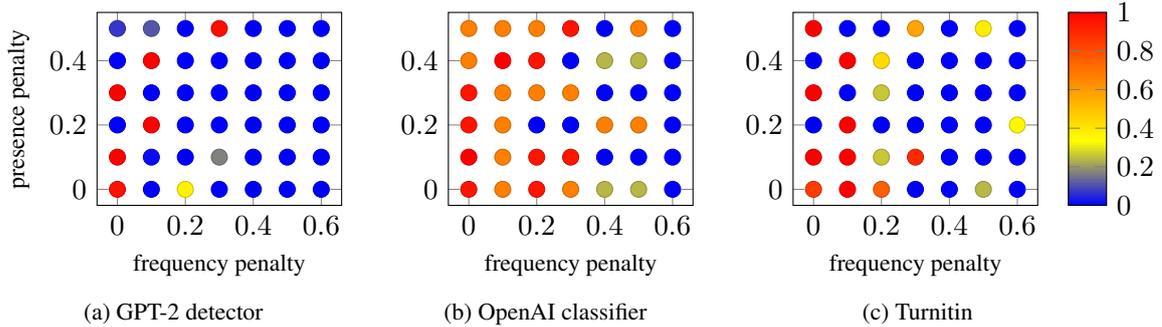
\begin{figure*}[ht!]
\vspace{-0.3cm}
     \centering
     \begin{subfigure}[b]{0.3\textwidth}
         \centering
             \begin{tikzpicture}
              \begin{axis}[width=\textwidth, xlabel={\small frequency penalty}, ylabel={\small presence penalty}]
                \addplot[opacity=1, mark size=3pt, scatter, only marks] table [x index=0, y index=1, scatter src=\thisrowno{3}, col sep=comma] {data/dataframe.csv};
              \end{axis}
            \end{tikzpicture}
         \caption{GPT-2 detector}
         \label{fig:y equals x}
     \end{subfigure}
     \begin{subfigure}[b]{0.3\textwidth}
         \centering
         \begin{tikzpicture}
              \begin{axis}[width=\textwidth, xlabel={\small frequency penalty}]
                \addplot[opacity=1, mark size=3pt, scatter, only marks] table [x index=0, y index=1, scatter src=\thisrowno{4}, col sep=comma] {data/dataframe.csv};
              \end{axis}
            \end{tikzpicture}
         \caption{OpenAI classifier}
         \label{fig:three sin x}
     \end{subfigure}
     \begin{subfigure}[b]{0.3\textwidth}
         \centering
         \begin{tikzpicture}
              \begin{axis}[width=\textwidth, xlabel={\small frequency penalty}, colorbar]
                \addplot[opacity=1, mark size=3pt, scatter, only marks] table [x index=0, y index=1, scatter src=\thisrowno{5}, col sep=comma] {data/dataframe.csv};
              \end{axis}
            \end{tikzpicture}
         \caption{Turnitin}
         \label{fig:five over x}
     \end{subfigure}
        \caption{Influence of joined optimsation of fequency and presence penalties on the detection score\vspace{-0.4cm}}
        \label{fig:interaction}
\end{figure*}

\subsection{Parameter tweaking}
First, we investigated the influence of GPT-3.5 generation parameters on the detection. Table \ref{tab:param} shows the parameters we focus on because they have the biggest impact on the produced texts according to \citet{openaiAPI}. Temperature and top p control randomness in the text. By increasing the temperature, the output becomes more random. However, for values beyond the default of 1.0, the length of the outputs started fluctuating strongly and the quality of the texts dropped. Consequently, we focused on the range between 0.0 and 1.0. Top p represents the percentage of tokens selected based on their probability mass. The frequency penalty controls the frequency of tokens appearing in the text, with higher values leading to more diverse verbatim. During the testing phase, it was found that increasing the frequency penalty beyond 1.0 degrades the quality of the texts rapidly. Additionally, as decreasing the value below 0.0 increases the repetitiveness, we focused on the range between 0.0 and 1.0. Finally, the presence penalty controls the model’s likelihood of repeating tokens in the text. Higher values of presence penalty lead to the model producing more diverse texts. Following consideration similar to the frequency penalty, the negative values were discarded, and we focused on the range between 0.0 and 2.0. \cite{openaiAPI} First, we investigated each parameter separately, changing it in steps of 0.1. Subsequently, we used the two parameters that had the biggest impact and investigated whether their interaction also influences the detection rate by performing a grid search in steps of 0.1 with these parameters.

\subsection{Prompt engineering}
The second approach explored the effect of prompt engineering on detection rates. Since \citet{Liang2023} have shown that detection is vulnerable towards simple rephrasing, our hypothesis was that providing regeneration instructions as a separate prompt might lower the detection rate. This hypothesis was tested with the following four prompts, each of which was used to generate ten texts: (1) standard prompt as described in Section; (2) 
standard prompt followed by second query \textit{"Regenerate the essay."}; (3) advanced prompt as shown in Appendix \ref{sec:advancedprompt}; (4) standard prompt followed by advanced prompt as second query. Additionally, several prompts to increase the perplexity and burstiness of generated texts were tested, a strategy that is popular among users (see e.g. \citet{perplexity}). Ten prompts were designed, and for each, ten texts were generated. The first five prompts use single-query architecture, while the others utilise a two-query concept. The following methods were tested with different prompts: (1) Explaining burstiness and perplexity, then asking to implement them (see Appendix \ref{sec:perplexity}); (2) Explicitly asking to maximise either perplexity, burstiness or both (standard prompt followed by \textit{``Maximize the burstiness / perplexity / burstiness and perplexity of the text.''}); (3) Explicitly asking to rewrite to avoid detection. (\textit{``Rewrite the above essay in order to avoid AI detection.''})

\subsection{Character-level mutations}
This approach aimed to test the robustness of the detectors against traditional adversarial attack vectors. Three character level mutations were taken as a basis: replacing either Latin lowercase “a” or “e” with the corresponding Cyrillic analogue; replacing Latin lowercase “L” with Latin uppercase “I”. Ten texts were generated with the standard prompt and then the mutations were applied.

\section{Results}

\subsection{Parameter tweaking}
The detection rate of all three detectors dropped with an increment in either frequency or presence penalty (see Figure \ref{fig:influence}). Starting from a frequency penalty of 0.3-0.4 and a presence penalty of 1.0-1.2, the detection rate fell under 50\% with some fluctuations. An analysis of the generated texts showed that increasing either the frequency or presence penalty led to more diverse texts. A higher frequency penalty caused a wider vocabulary variety. For values above 0.6, the occurrence of punctuation mistakes and unclear wordings quickly increased, making the texts difficult to read. For values between 0.0 and 0.6, increases in value caused an incremental increase in text complexity while preserving quality and readability. A higher presence penalty primarily influenced the diversity of perspectives and text engagement. However, for values above 0.6, the coherence and logical progression rapidly decreased and texts became less subject-focused. Therefore, increasing either frequency or presence penalty from the default of 0.0 up to 0.6 can be seen as successful attack strategies that significantly lowers detection while maintaining text quality. Increasing the temperature and top p value above the default was already excluded from the experimental design, because of the strong negative effects on text quality. Lowering either value resulted in more deterministic outputs leading to higher detection rates, therefore, tweaking those parameters is not a successful attack strategy.

In a second step, the interaction between frequency and presence penalty was investigated. Figure \ref{fig:interaction} shows that the detection rates dropped for all three detectors with an increment in frequency and presence penalty. Notably, they started dropping for smaller values of presence and frequency penalty than they did for the individual parameters, thereby minimising the potential negative effects on text quality. The GPT-2 detector showed the worst performance in the comparison with very quickly declining detection scores.

\subsection{Prompt engineering}

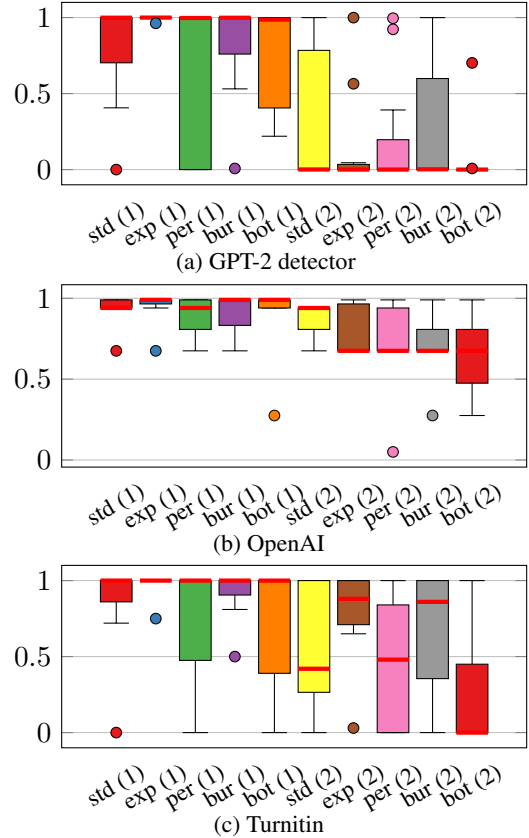
\begin{figure}[t]
    \centering
    \begin{subfigure}[b]{\linewidth}
    \centering
    \begin{tikzpicture}
	\pgfplotstableread[col sep=comma]{data/gpt2_scores.csv}\csvdata
	\pgfplotstabletranspose\datatransposed{\csvdata} 
	\begin{axis}[
            height=4cm,
            width=\linewidth,
		boxplot/draw direction = y,
		ymajorgrids,
		xtick = {1, 2, 3, 4,5,6,7,8,9,10},
		xticklabel style = {align=center, font=\small, rotate=30},
		xticklabels = {std (1), exp (1), per (1), bur (1), bot (1), std (2), exp (2), per (2), bur (2), bot (2)},
		xtick style = {draw=none}, 
		ytick = {0, 0.5,1},
            boxplot/every median/.style={red,ultra thick}
	]
		\foreach \n in {1,...,10} {
			\addplot+[boxplot, fill, draw=black, mark = *] table[y index=\n] {\datatransposed};
		}
	\end{axis}
\end{tikzpicture}
\vspace{-0.4cm}
    \caption{GPT-2 detector
\label{fig:bursgpt2}}        
    \end{subfigure}
    \begin{subfigure}[b]{\linewidth}
    \centering
    \begin{tikzpicture}
	\pgfplotstableread[col sep=comma]{data/openai_scores.csv}\csvdata
	\pgfplotstabletranspose\datatransposed{\csvdata} 
	\begin{axis}[
            height=4cm,
            width=\linewidth,
		boxplot/draw direction = y,
		ymajorgrids,
		xtick = {1, 2, 3, 4,5,6,7,8,9,10},
		xticklabel style = {align=center, font=\small, rotate=30},
		xticklabels = {std (1), exp (1), per (1), bur (1), bot (1), std (2), exp (2), per (2), bur (2), bot (2)},
		xtick style = {draw=none}, 
		ytick = {0, 0.5,1},
            boxplot/every median/.style={red,ultra thick}
	]
		\foreach \n in {1,...,10} {
			\addplot+[boxplot, fill, draw=black, mark = *] table[y index=\n] {\datatransposed};
		}
	\end{axis}
\end{tikzpicture}
\vspace{-0.4cm}
    \caption{OpenAI}        
    \end{subfigure}
    \begin{subfigure}[b]{\linewidth}
    \centering
    \begin{tikzpicture}
	\pgfplotstableread[col sep=comma]{data/turnitin_scores.csv}\csvdata
	\pgfplotstabletranspose\datatransposed{\csvdata} 
	\begin{axis}[
            height=4cm,
            width=\linewidth,
		boxplot/draw direction = y,
		ymajorgrids,
		xtick = {1, 2, 3, 4,5,6,7,8,9,10},
		xticklabel style = {align=center, font=\small, rotate=30},
		xticklabels = {std (1), exp (1), per (1), bur (1), bot (1), std (2), exp (2), per (2), bur (2), bot (2)},
		xtick style = {draw=none}, 
		ytick = {0, 0.5,1},
            boxplot/every median/.style={red,ultra thick}
	]
		\foreach \n in {1,...,10} {
			\addplot+[boxplot, fill, draw=black, mark = *] table[y index=\n] {\datatransposed};
		}
	\end{axis}
\end{tikzpicture}
\vspace{-0.4cm}
    \caption{Turnitin}        
    \end{subfigure}
   \caption{Influence of perplexity and burstiness prompts on detection (number of prompts in brackets; std = standard prompt; exp = prompt to explain and increase burstiness \& perplexity; per = increase perplexity; bur = increase burstiness; bot = increase both)\vspace{-0.4cm}}
    \label{fig:burst}
\end{figure}

The simple regeneration approach, whether using a second query or providing detailed instruction in the first query, did not have an impact on detection rates. Increasing the perplexity and burstiness of the texts through prompts, on the other hand, caused a drop in the detection rate across all three detectors, however only if applied in two separate prompts, as shown in Figure \ref{fig:burst}. Although the approach managed to decrease the detection rate across all three detectors, the score sank only for the GPT-2 detector below 0.5 (see Figure \ref{fig:bursgpt2}). For the other two detectors, the score stayed above 0.5, meaning that the generated texts would still be detected as AI-generated or at least as undecidable.

\subsection{Character-level mutations}
The influence of the character-level mutations on the detection scores is shown in Table \ref{tab:mutation}. For the GPT-2 detector, all three replacements lead to a detection score of 0. For Open AI, all attacks lowered the detection score, although not as much as for the GPT-2 detector. Turnitin detected the replacement of Latin characters with Cyrillic characters and flagged the attack. The substitution of \textit{l}s with capital \textit{i}s, however, remained undetected and significantly lowered the detection score, therefore presenting a successful attack strategy.

\begin{table}[b!]
\centering
\begin{tabular}{lrrr}
\hline
{} &  GPT-2 &  OpenAI & Turnitin \\\hline
Standard                       &    0.67 &   0.77 &     0.75 \\
Swap a lat.-cyr.       &     0 &      0.52 &     x \\
Swap e lat.-cyr.      &     0 &      0.48 &     x \\
Swap l - I &     0 &      0.38 &      0.21 \\
\hline
\end{tabular}
\caption{Character-level mutation mean detection score}
\label{tab:mutation}
\end{table}

\section{Conclusion}
This study explored the effectiveness of resource-efficient adversarial attacks on neural text detectors, based on texts generated by GPT-3.5 and the three detectors GPT-2 output detector, OpenAI classifier, and Turnitin. Of the three investigated strategies, parameter tweaking and character-level mutations were successful for all three detectors. Prompt engineering was only successful for the GPT-2 output detector. All strategies are resource efficient and easy to implement, effectively showing that currently available detectors cannot reliably detect AI-generated texts and are vulnerable to adversarial attacks.

\bibliography{anthology,custom}

\begin{thebibliography}{21}
\expandafter\ifx\csname natexlab\endcsname\relax\def\natexlab#1{#1}\fi

\bibitem[{Alexander(2023)}]{perplexity}
Chris Alexander. 2023.
\newblock Asking chatgpt to put perplexity and burstiness in an essay appears
  to fool ai detectors.
\newblock
  \url{https://telblog.unic.ac.cy/teaching-chatgpt-perplexity-burstiness}
  \url{-appears-to-fool-ai-detectors/}. Last accessed: 2023-07-11.

\bibitem[{Crothers et~al.(2022)Crothers, Japkowicz, Viktor, and
  Branco}]{Crothers2022}
Evan Crothers, Nathalie Japkowicz, Herna Viktor, and Paula Branco. 2022.
\newblock \href {https://doi.org/10.1109/IJCNN55064.2022.9892269} {Adversarial
  robustness of neural-statistical features in detection of generative
  transformers}.
\newblock \emph{Proceedings of the International Joint Conference on Neural
  Networks}, 2022-July.

\bibitem[{Ebrahimi et~al.(2018)Ebrahimi, Rao, Lowd, and
  Dou}]{ebrahimi-etal-2018-hotflip}
Javid Ebrahimi, Anyi Rao, Daniel Lowd, and Dejing Dou. 2018.
\newblock \href {https://doi.org/10.18653/v1/P18-2006} {{H}ot{F}lip: White-box
  adversarial examples for text classification}.
\newblock In \emph{Proceedings of the 56th Annual Meeting of the Association
  for Computational Linguistics (Volume 2: Short Papers)}, pages 31--36,
  Melbourne, Australia. Association for Computational Linguistics.

\bibitem[{Gao et~al.(2018)Gao, Lanchantin, Soffa, and Qi}]{Gao2018}
Ji~Gao, Jack Lanchantin, Mary~Lou Soffa, and Yanjun Qi. 2018.
\newblock \href {https://doi.org/10.1109/SPW.2018.00016} {Black-box generation
  of adversarial text sequences to evade deep learning classifiers}.
\newblock \emph{Proceedings - 2018 IEEE Symposium on Security and Privacy
  Workshops, SPW 2018}, pages 50--56.

\bibitem[{Goodfellow et~al.(2014)Goodfellow, Shlens, and
  Szegedy}]{Goodfellow2014}
Ian~J. Goodfellow, Jonathon Shlens, and Christian Szegedy. 2014.
\newblock \href {https://arxiv-org.ezproxy2.utwente.nl/abs/1412.6572v3}
  {Explaining and harnessing adversarial examples}.
\newblock \emph{3rd International Conference on Learning Representations, ICLR
  2015 - Conference Track Proceedings}.

\bibitem[{Jawahar et~al.(2020)Jawahar, Abdul-Mageed, and
  Lakshmanan}]{jawahar-etal-2020-automatic}
Ganesh Jawahar, Muhammad Abdul-Mageed, and Laks Lakshmanan, V.S. 2020.
\newblock \href {https://doi.org/10.18653/v1/2020.coling-main.208} {Automatic
  detection of machine generated text: A critical survey}.
\newblock In \emph{Proceedings of the 28th International Conference on
  Computational Linguistics}, pages 2296--2309, Barcelona, Spain (Online).
  International Committee on Computational Linguistics.

\bibitem[{Jin et~al.(2019)Jin, Jin, Zhou, and Szolovits}]{Jin2019}
Di~Jin, Zhijing Jin, Joey~Tianyi Zhou, and Peter Szolovits. 2019.
\newblock \href {https://doi.org/10.1609/aaai.v34i05.6311} {Is bert really
  robust? a strong baseline for natural language attack on text classification
  and entailment}.
\newblock \emph{AAAI 2020 - 34th AAAI Conference on Artificial Intelligence},
  pages 8018--8025.

\bibitem[{Kirchenbauer et~al.(2023)Kirchenbauer, Geiping, Wen, Katz, Miers, and
  Goldstein}]{Kirchenbauer2023}
John Kirchenbauer, Jonas Geiping, Yuxin Wen, Jonathan Katz, Ian Miers, and Tom
  Goldstein. 2023.
\newblock \href {https://arxiv-org.ezproxy2.utwente.nl/abs/2301.10226v3} {A
  watermark for large language models}.

\bibitem[{Liang et~al.(2023{\natexlab{a}})Liang, Guerrero, and
  Alsmadi}]{liang2023mutationbased}
Gongbo Liang, Jesus Guerrero, and Izzat Alsmadi. 2023{\natexlab{a}}.
\newblock \href {http://arxiv.org/abs/2302.05794} {Mutation-based adversarial
  attacks on neural text detectors}.

\bibitem[{Liang et~al.(2023{\natexlab{b}})Liang, Guerrero, Zheng, and
  Alsmadi}]{electronics12081948}
Gongbo Liang, Jesus Guerrero, Fengbo Zheng, and Izzat Alsmadi.
  2023{\natexlab{b}}.
\newblock \href {https://doi.org/10.3390/electronics12081948} {Enhancing neural
  text detector robustness with \&mu;attacking and rr-training}.
\newblock \emph{Electronics}, 12(8).

\bibitem[{Liang et~al.(2023{\natexlab{c}})Liang, Yuksekgonul, Mao, Wu, and
  Zou}]{Liang2023}
Weixin Liang, Mert Yuksekgonul, Yining Mao, Eric Wu, and James Zou.
  2023{\natexlab{c}}.
\newblock \href {https://openreview.net/forum?id=SPuX8tKKIQ} {{GPT} detectors
  are biased against non-native english writers}.
\newblock In \emph{ICLR 2023 Workshop on Trustworthy and Reliable Large-Scale
  Machine Learning Models}.

\bibitem[{Nova(2019)}]{essayTopics}
A.~Nova. 2019.
\newblock Essay topics: 100+ best essay topics for your guidance.
\newblock
  \url{https://www.5staressays.com/blog/essay-writing-guide/essay-topics}.
\newblock Last accessed: 2023-07-11.

\bibitem[{OpenAI(2023{\natexlab{a}})}]{openaiClassifier}
OpenAI. 2023{\natexlab{a}}.
\newblock Ai text classifier - openai api.
\newblock \url{https://platform.openai.com/ai-text-classifier}.
\newblock Last accessed: 2023-07-11.

\bibitem[{OpenAI(2023{\natexlab{b}})}]{openaiAPI}
OpenAI. 2023{\natexlab{b}}.
\newblock \href {https://platform.openai.com/docs/api-reference/chat/create}
  {Api reference - openai api}.

\bibitem[{Peng et~al.(2023)Peng, Wang, Zhao, Wu, Han, Guo, Ji, and
  Zhong}]{Peng2023}
Hao Peng, Zhe Wang, Dandan Zhao, Yiming Wu, Jianming Han, Shixin Guo, Shouling
  Ji, and Ming Zhong. 2023.
\newblock \href {https://doi.org/10.1016/J.JKSUCI.2023.03.017} {Efficient
  text-based evolution algorithm to hard-label adversarial attacks on text}.
\newblock \emph{Journal of King Saud University - Computer and Information
  Sciences}, 35:101539.

\bibitem[{Rathore et~al.(2021)Rathore, Basak, Nistala, and
  Runkana}]{Rathore2021}
Pradeep Rathore, Arghya Basak, Sri~Harsha Nistala, and Venkataramana Runkana.
  2021.
\newblock \href {https://doi.org/10.1109/IJCNN48605.2020.9207272} {Untargeted,
  targeted and universal adversarial attacks and defenses on time series}.
\newblock \emph{Proceedings of the International Joint Conference on Neural
  Networks}.

\bibitem[{Sadasivan et~al.(2023)Sadasivan, Kumar, Balasubramanian, Wang, and
  Feizi}]{Sadasivan2023}
Vinu~Sankar Sadasivan, Aounon Kumar, Sriram Balasubramanian, Wenxiao Wang, and
  Soheil Feizi. 2023.
\newblock \href {http://arxiv.org/abs/2303.11156} {Can ai-generated text be
  reliably detected?}

\bibitem[{Shen et~al.(2023)Shen, Zhang, Ji, Pu, Ge, Yang, and Feng}]{Shen2023}
Lujia Shen, Xuhong Zhang, Shouling Ji, Yuwen Pu, Chunpeng Ge, Xing Yang, and
  Yanghe Feng. 2023.
\newblock \href {http://arxiv.org/abs/2302.05892} {Textdefense: Adversarial
  text detection based on word importance entropy}.

\bibitem[{Szegedy et~al.(2013)Szegedy, Zaremba, Sutskever, Bruna, Erhan,
  Goodfellow, and Fergus}]{Szegedy2013}
Christian Szegedy, Wojciech Zaremba, Ilya Sutskever, Joan Bruna, Dumitru Erhan,
  Ian Goodfellow, and Rob Fergus. 2013.
\newblock \href {https://arxiv-org.ezproxy2.utwente.nl/abs/1312.6199v4}
  {Intriguing properties of neural networks}.
\newblock \emph{2nd International Conference on Learning Representations, ICLR
  2014 - Conference Track Proceedings}.

\bibitem[{Wolff and Wolff(2022)}]{wolff2022attacking}
Max Wolff and Stuart Wolff. 2022.
\newblock \href {http://arxiv.org/abs/2002.11768} {Attacking neural text
  detectors}.

\bibitem[{Yoo et~al.(2022)Yoo, Kim, Jang, and Kwak}]{Yoo2022}
Ki~Yoon Yoo, Jangho Kim, Jiho Jang, and Nojun Kwak. 2022.
\newblock \href {https://doi.org/10.18653/v1/2022.findings-acl.289} {Detection
  of word adversarial examples in text classification: Benchmark and baseline
  via robust density estimation}.
\newblock \emph{Proceedings of the Annual Meeting of the Association for
  Computational Linguistics}, pages 3656--3672.

\end{thebibliography}
\bibliographystyle{acl_natbib}
\newpage
\appendix

\section{Appendix - Prompts}
\label{sec:appendix}

\subsection{Advanced Prompt}
\label{sec:advancedprompt}
\textit{``Write a five-hundred-word argumentative essay on the topic ‘topic’. Include personal reflections, use a mix of long and short sentences, employ rhetorical questions to engage the reader, maintain a conversational tone in parts, and play around with the paragraph structure to create a dynamic and engaging piece of writing. Try to include factual and contextual information, use advanced concepts and vocabulary. Utilize a combination of complex and simple vocabulary. Try to mimic human writing as closely as you can. Avoid passive voice, as it tends to occur more often in AI-generated texts. Add a few examples from the real world illustrating your point.''}

\subsection{Perplexity and Burstiness}
\label{sec:perplexity}
\textit{``Write a five-hundred-word argumentative essay on the topic ‘topic’. When it comes to writing content, two factors are crucial, perplexity and burstiness. Perplexity measures the complexity of the text. Separately, burstiness compares the variations of sentences. Humans tend to write with greater burstiness, for example, with some longer or more complex sentences alongside shorter ones. AI sentences tend to be more uniform. Therefore, when writing the following content, I need it to have a good amount of perplexity and burstiness.}

\end{document}